%% file: main.tex
\documentclass[runningheads]{llncs}
\usepackage{graphicx}
\usepackage{amsmath}
\usepackage{multirow}
\usepackage{amssymb}
\newcommand*\samethanks[1][\value{footnote}]{\footnotemark[#1]}
\begin{document}

\title{FairLens: Auditing Black-box \\Clinical Decision Support Systems}
%
%
\author{Cecilia Panigutti\inst{1} \and
Alan Perotti\inst{2}\thanks{PB, AP and AP acknowledge partial support from Research Project "Casa Nel Parco" (POR FESR14/20 - CANP - Cod. 320 - 16 - Piattaforma Tecnologica "Salute e Benessere") fundedby Regione Piemonte in the context of the Regional Platform on Health and Wellbeing and from Intesa Sanpaolo Innovation Center. The funders had no role in study design,data collection and analysis, decision to publish, or preparation of the manuscript}\and
Andrè Panisson\inst{2}\samethanks\\ \and
Paolo Bajardi\inst{2}\samethanks\and
Dino Pedreschi\inst{3}}
\authorrunning{C. Panigutti et al.}
%
\institute{Scuola Normale Superiore \email{name.surname@sns.it} \and
ISI Foundation  \email{name.surname@isi.it}\and
University of Pisa  \email{name.surname@unipi.it}}
\maketitle              

\begin{abstract}
The pervasive application of algorithmic decision-making is raising concerns on the risk of unintended bias in AI systems deployed in critical settings such as healthcare. The detection and mitigation of biased models is a very delicate task which should be tackled with care and involving domain experts in the loop. In this paper we introduce FairLens, a methodology for discovering and explaining biases. We show how our tool can be used to audit a fictional commercial black-box model acting as a clinical decision support system. In this scenario, the healthcare facility experts can use FairLens on their own historical data to discover the model's biases before incorporating it into the clinical decision flow. FairLens first stratifies the available patient data according to attributes such as age, ethnicity, gender and insurance; it then assesses the model performance on such subgroups of patients identifying those in need of expert evaluation. Finally, building on recent state-of-the-art XAI (eXplainable Artificial Intelligence) techniques, FairLens explains which elements in patients' clinical history drive the model error in the selected subgroup. Therefore, FairLens allows experts to investigate whether to trust the model and to spotlight group-specific biases that might constitute potential fairness issues.


\end{abstract}

\pagebreak

\input{01_intro.tex}
\input{02_bg_rw.tex}

\input{03_pipeline.tex}
\input{04_experiments.tex}
\input{05_conclusions.tex}


\bibliographystyle{splncs04}
\bibliography{the_bibliography.bib}

\end{document}

%% file: 01_intro.tex
\section{Introduction}

The growing availability of Electronic Health Records (EHR) and the constantly increasing predictive power of Machine Learning (ML) models are boosting both research advances and the creation of business opportunities to deploy clinical decision support systems (DSS) in healthcare facilities~\cite{jiang2017artificial,davenport2019potential,moja2019effectiveness}. Since such models are still not equipped to differentiate between correlation and causation, they might leverage spurious correlations and undesired biases to boost their performance. While there is an increasing interest of the AI community to commit to interdisciplinary endeavors to define, investigate and provide guidelines to tackle biases and fairness-related issues~\cite{pedreschi2008discrimination,saleiro2018aequitas,obermeyer2019dissecting}, quantitative and systematic auditing for real-world datasets and ML models is still in its infancy. In this work, we investigate the potential biases in ML models trained on patients' clinical history represented as diagnostic codes. This type of structured data allows for a machine-ready representation of the patient's clinical history and is commonly used in longitudinal ML modeling for phenotyping, multi-morbidity diagnosis classification and sequential clinical events prediction~\cite{xiao2018opportunities,choi2016doctor,che2018recurrent}. The implicit assumption behind the use of ICD codes in this kind of ML applications is that these codes are a good proxy for the patient's actual health status. However, ICD codes can misrepresent such status because of many potential sources of error in translating the patient's actual disease into the respective code~\cite{o2005measuring,chen2019can}. Data quality assessment for the secondary use of health data is of pivotal importance to ease the transition of ML-based DSS from academic prototypes into real-world clinical practice. This is particularly true when ICD codes are fed into \emph{black-box} ML models, i.e., models whose internal decision-making process is opaque. \\

In this paper we introduce FairLens, a methodology to discover and explain biases for ML models trained on ICD structured healthcare data. We take bias analysis a step further by explaining the reasons behind the bad model's performance on specific subgroups. FairLens embeds explainability techniques in order to explain the reasons behind model mistakes instead of simply explaining model predictions. The presented methodology is designed to be applied to any sequential ML model trained on ICD codes. FairLens first stratifies patients according to attributes of interest such as age, gender, ethnicity, and insurance type; it then applies an appropriate metric to identify patients subgroups where the model performs poorly. Lastly, FairLens identifies the clinical conditions that are most frequently misclassified for the selected subgroup and explains which elements in the patients' clinical histories are influencing the misclassification. We also present a use case for our methodology using the most recent update of one of the largest freely available ICU datasets, the MIMIC-IV dataset~\cite{johnson2020mimiciv}. In this scenario, MIMIC-IV acts as the healthcare facility's historical medical database. We show how a domain expert can use FairLens to audit a multilabel clinical DSS~\cite{choi2016doctor} acting as a fictional commercial black-box model.  We believe that applied research and quantitative tools to perform systematic audits specific to healthcare data are very much needed in order to establish and reinforce trust in the application of AI-based systems in such a high-stakes domain. FairLens is a first step to make fairness and bias auditing a standard procedure for clinical DSS. We envision such a procedure to be used to monitor bias and fairness issues in all clinical DSSs' life-cycle stages.

%% file: 02_bg_rw.tex
\section{Background and Related Work}\label{s:bgrw}




Advances in artificial intelligence (AI) in healthcare offer groundbreaking opportunities to enhance patient outcomes, reduce costs, and impact population health \cite{yu2018artificial,topol2019high}. Unprecedented results have been achieved leveraging deep neural networks for pattern recognition to help interpret medical scans~\cite{lindsey2018deep,nam2019development,chilamkurthy2018deep,titano2018automated}, pathology slides~\cite{bejnordi2017diagnostic,coudray2018classification,capper2018dna}, skin lesions \cite{esteva2017dermatologist,haenssle2018man}, retinal images \cite{gulshan2016development,abramoff2018pivotal} and electrocardiograms \cite{madani2018fast,zhang2018fully} to name few examples. The ability to predict key outcomes can also be exploited to improve clinical practice by training DSSs with electronic health records~\cite{rajkomar2018scalable,avati2018improving,shameer2017predictive,norgeot2019call,chen2017disease}.\\

Besides the technological challenges, the various stakeholders involved with the healthcare ecosystem (clinicians, patients/ patient advocate, researchers, federal agencies and industry) identified the following urgent priorities for healthcare applications: trustworthiness, explainability, usability, transparency and fairness \cite{cutillo2020machine}.
Many efforts have been devoted to detecting and measuring discrimination in model decisions \cite{ruggieri2010data,zemel2013learning,hajian2016algorithmic}.
Several definitions and methodologies have been proposed to measure bias and fairness \cite{pedreschi2008discrimination,dwork2012fairness,luong2011k,hardt2016equality}; however, despite the effort, a general consensus on such measures is still missing. Generally speaking, the most prevalent approach to fairness in machine learning is to solicit for approximate parity of some statistics of the predictions (such as false negative rate) across pre-defined subgroups \cite{kleinberg2016inherent,kearns2018preventing,chouldechova2017fair}. Moreover, there are very few available general-purposes resources to operationalize them \cite{adebayo2016fairml,tramer2017fairtest,bellamy2018ai,saleiro2018aequitas}.
The majority of such research has focused on binary or multi-class classification problems to prevent discrimination based on sensitive attributes
\cite{feldman2015certifying}, and a few studies focus specifically on multi-label classification problems,
with many concentrating on fairness in ranking  and recommendation systems \cite{abdollahpouri2017controlling,edizel2020fairecsys}.
In the context of medical applications, a recent paper \cite{chen2020ethical} suggested that the post-deployment inspection of model performance on groups and outcomes should be one out of five ethical pillars for equitable ML in the advancement of health care.\\

Another staple of this paper is the research field of  eXplainable Artificial Intelligence (XAI)~\cite{gunning2017explainable}. XAI techniques have the goal to \emph{explain} (i.e., present in human-understandable terms) the decision-making process of an AI system.
The need for this kind of technique stems from the fact that the internal decision-making process of many state-of-the-art AI systems is opaque.
This can happen either because the source code of the algorithm is proprietary software and can not be directly inspected, or because the model implements a subsymbolic (numerical) representation of knowledge, often paired with highly non-linear correlations, or both. 
Recently, several state-of-the-art XAI techniques have been proposed to provide post-hoc explanations of the model behaviour~\cite{ribeiro2016should,lundberg2017unified,guidotti2018survey,panigutti2020doctor}. The post-hoc approach to model explanation have recently faced criticism~\cite{rudin2019stop}, however they remain the only available solution in case of proprietary software.

%% file: 03_pipeline.tex
\section{FairLens: Pipeline}\label{s:pipeline}

\begin{figure}[ht!]
  \includegraphics[width=\textwidth]{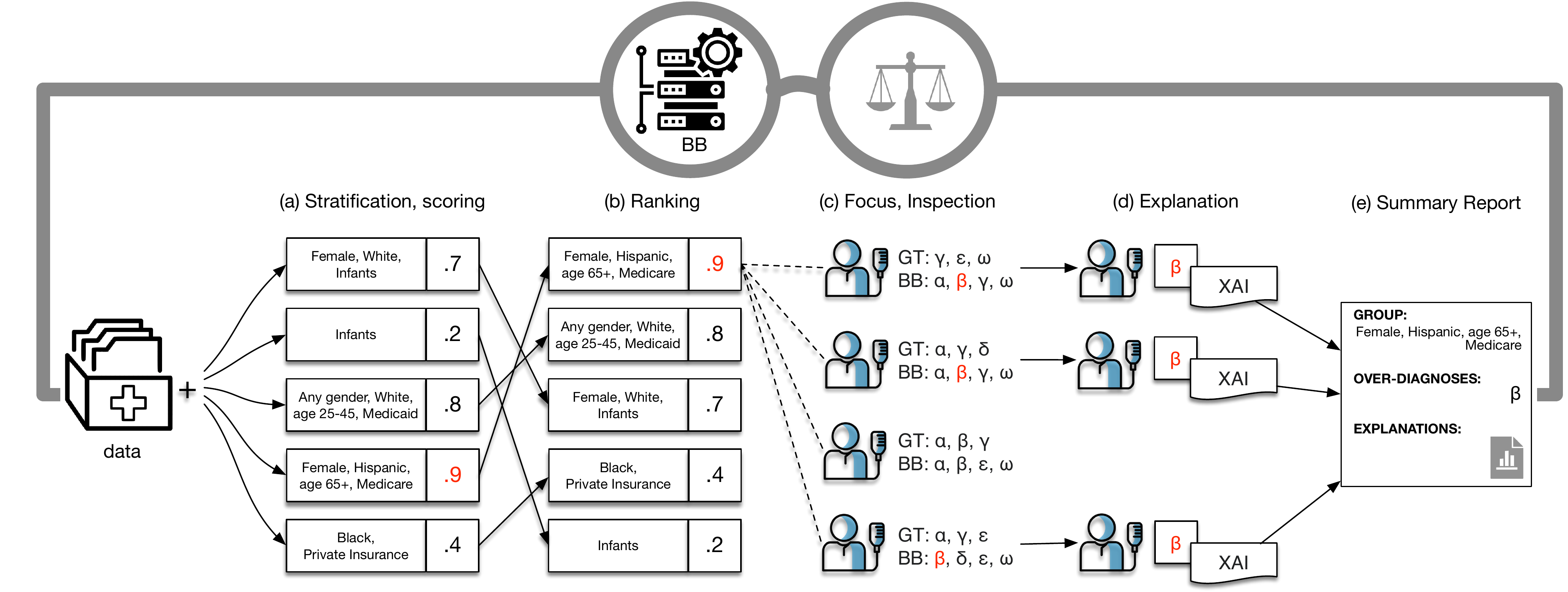}
  \caption{FairLens pipeline}
  \label{fig:pipeline}
\end{figure}

This Section describes the FairLens methodology. A bird's-eye view of the pipeline is depicted in Figure~\ref{fig:pipeline}.

Let $BB$ be a sequential black-box ML model trained on ICD data. The model can be available as an on-premise-installed software or it could be integrated via SaaS (software as a service).

Let $p_i = (p_i^{att},p_i^{ch})$ be the patients represented by a set of attributes $p_i^{att}$ such as \emph{ethnicity}, \emph{gender}, and \emph{insurance type}, and by a clinical history $p_i^{ch} = \{v_{i,1}, \cdots v_{i,V}\}$ represented as a sequence of visits. In turn, each visit is represented by a set of ICD codes. Let $P = \{p_1, \cdots p_N\}$ be the set of patients.

Let $v_{i,j}^{BB} = BB(\{v_{i,1}, \cdots v_{i,j-1} \})$ be the prediction of the black-box for the $j-th$ visit of patient $p_i$. \\


\textbf{Stratification} The first step of our methodology is depicted in Figure~\ref{fig:pipeline}(a).

Since 
we aim to check
whether the ML model performs equally well across groups, we stratify our patients set $P$ according to a set of conditions $c$ on the set of attributes $p^{att}$, e.g. $c = \{ \text{age} \leq 40, \text{insurance} = \text{Medicaid} \}$. We define a {\em group} $G$ as the set of visits of each patients whose attributes match the conditions in $s$:
$$G_j = \{p_i \mid p_i \in P, p_i^{att} \in c_j\}$$
The stratification process produces a set of groups $G_1,..,G_M$. The black-box model is then applied on such patients obtaining a prediction for each of the selected visits. 


A domain expert might suggest specific condition sets to isolate a given sub-cohort of known interest, whereas a technician might opt for building a lattice of all possible combinations of constraints. We also remark that some patients might not occur in any group or occur in more than one, depending on the provided conditions.\\ 

\textbf{Performance evaluation and ranking} A \textit{disparity function} $d:G_j \to s_j$ maps every group $G_j$ to a \textit{disparity score} $s_j$.
FairLens includes a number of disparity functions, such as the standard classification metrics (accuracy, F1-score, etc.) and distribution-comparison functions like the Wasserstein distance. 
Custom disparity functions can be used, as long as 
their results can be used for ranking.
Given a disparity function, FairLens computes the score $s_j$ for each group $G_j$, which represents the performance of the $BB$ on that specific set of patients. Once each group has been scored, FairLens ranks the groups, as depicted in Figure~\ref{fig:pipeline}(b). 
The ranking highlights groups where the $BB$ performs relatively poorly, signaling them to domain experts for further inspection. Alternatively, the domain experts might arbitrarily select one subgroup for further inspection, regardless of their scores, due to the cohort's known peculiarities or clinical-dependent reasons.\\


\textbf{Inspection} Given a specific group $G_j$ flagged for further inspection by the group ranking function, FairLens compares the black-box prediction $v_{i,j}^{BB}$ with the ground truth $v_{i,j}$ for each visit in $G_j$. The goal of this step is to check for systematic bias of the $BB$ on the group of patients. For each ICD code, the relative frequencies in the predicted and true values are computed and we define the \textit{misdiagnosis score} the difference between these two values. Ranking the codes by misdiagnosis scores allows to highlight which ICD codes are particularly over- or under-predicted (high and low difference values respectively). FairLens thus displays the top three over- and under-represented codes to the domain expert who can ask for an explanation for the highlighted conditions that might result in producing or reinforcing systematic over- or under-treatment. In Figure~\ref{fig:pipeline}(c), we have labelled the true visit value as $GT$ (for {\em ground truth}); in the mock example it can be observed that the code $\beta$ is over-represented. \\


\textbf{Explanation} In order to extract an explanation for the mislabeled code, FairLens first assigns binary labels on the visits of the group of interest. Suppose the domain expert wants to understand what elements of the group clinical histories are most influencing the over-representation of ICD code $\beta$ in the inspected group $G_j$, then at each visit $v_{i,j} \in G_j$ will be assigned a binary label in the following way:

\[ l(v_{i,j}^{BB}) =
  \begin{cases}
    1  & \quad \text{if } \beta \in v_{i,j}^{BB}\\
    0  & \quad \text{if } \beta \notin v_{i,j}^{BB}
  \end{cases}
\]

By doing so, FairLens assigns to the clinical history $p_i^{ch} = {v_{i,1}, \cdots, v_{i,j}}$ of each patient a binary label representing patient misclassification w.r.t $\beta$. Then, it selects all the misclassified clinical histories and explains them using a local XAI technique for sequential healthcare data.
It is worth noting that while this kind of techniques are usually employed to explain the reasons behind a black-box decision, thanks to the binarisation process, FairLens uses them to explain the reasons behind a specific mislabelling. Furthermore, the use of a post-hoc XAI technique that is agnostic w.r.t. the black-box allows FairLens to audit any model without having access to its internal structure or parameters.\\


\textbf{Reporting} Finally, FairLens combines the explanations of each mislabelled patient in the group and shows the most common ICD codes occurring as patients' explanations. 
In this context, several local explanations of mislabelled visits related to patients of the same group are aggregated to provide the domain expert with a pointer to specific clinical conditions that systematically drive the $BB$ error. Typically the local model-agnostic post-hoc explanations provides local decision rules. Each condition of the rule premise follows the pattern
\begin{align*} \text{ICD}\_ \text{code} \gtrless threshold\_value\end{align*}
Where the \emph{threshold value} contains a temporal information about the visits containing the ICD code, being past clinical conditions predictive of future outcomes. In the aggregation process FairLens aggregate the occurrences of each ICD found in the selected group explanations and highlight the most common ones.

%% file: 04_experiments.tex
\section{Use Case: auditing a medical decision support system}\label{s:usecase}

In this section we show how a domain expert can use FairLens on the historical data available at her healthcare facility to audit a fictional commercial clinical \emph{decision support system} (DSS). We imagine that the domain expert has no access to the source code of the DSS, i.e. it can be considered a \emph{black-box}. We use the MIMIC-IV (see Subsection~\ref{ss:mimic}) database of electronic health records as the fictitious historical database of the facility and DoctorAI (see Subsection~\ref{ss:drai}) as the fictional clinical DSS. We split the dataset in training (29276 patients, 67\%), validation (4759 patients, 11\%) and test set (9662, 22\%). We used the training and validation set to train DoctorAI and the patients in the test set as the historical database used in the auditing process. We exploit DoctorXAI (Subsection~\ref{ss:drxai}) as the backbone of our explainer.


\subsection{Dataset: MIMIC-IV}\label{ss:mimic}

The MIMIC (Medical Information Mart for Intensive Care)~\cite{goldberger2000physiobank,johnson2016mimic} database is a single-center freely available database containing de-identified clinical data of patients admitted to the ICU (intensive care unit) of the Beth Israel Deaconess Medical Center in Boston. Its most recent update, MIMIC-IV~\cite{johnson2020mimiciv}, contains information of 383,220 patients collected between 2008 and 2019 for a total of 524,520 hospital admissions. The database includes patient's demographics, clinical measurements and diagnoses and procedures codes of each admission. We focused our analysis on hospital admissions coded with ICD-9 billing codes and on patients having at least two admissions to the hospital, reducing the number of patients to 43,697 and the number of admission to the hospital to 164,411 (see table~\ref{tab1}).

\begin{table}[h!]
\centering
\begin{small}
\begin{tabular}{|l|r|}
\cline{1-2}
number of patients                    & 43,697   \\ \cline{1-2}
number of admissions                  & 164,411  \\ \cline{1-2}
avg. nr. of admissions per patient & 3.76    \\ \cline{1-2}
max nr. of admissions per patient  & 146     \\ 
\cline{1-2}
number of unique ICD-9 codes          & 8,259    \\ 
\cline{1-2}
avg. nr. of codes per admission & 11.22  \\
\cline{1-2}
\end{tabular}
\end{small}
\vspace*{2mm}
\caption{M{\small IMIC-IV}: Data from patients with at least two hospital admissions}\label{tab1}
\end{table}




\subsection{Clinical DSS: Doctor AI}\label{ss:drai}

Doctor AI by~\cite{choi2016doctor} is a Recurrent Neural Network (RNN) with Gated Recurrent Units (GRU) that predicts the patient's next clinical event's time, diagnoses and medications. For the purpose of the use-case, we focused only on ICD diagnoses prediction. We trained the algorithm on MIMIC-IV using the training and validation set as defined in Subsection \ref{ss:mimic} using default hyperparameters.

Doctor AI can be trained to predict patient's future clinical event in terms of either CCS (Clinical Classifications Software) or ICD codes. CCS codes are used to group ICD codes into smaller number of clinically meaningful categories. As suggested in~\cite{choi2016doctor} we trained Doctor AI to predict CCS codes.

In order to test its performance we used Recall@$n$
with $n = 10, 20, 30$. Originally, the authors trained their model on 260.000 patients of the EHRs database of Sutter Health Palo Alto Medical Foundation. We consider the performance declared in the original paper as the DSS performance declared by the manufacturer. We show the performance on the original dataset compared to those on MIMIC-IV in table~\ref{tab:perf}. We highlight how this is the first step that a domain expert would do to check if the clinical DSS is suitable on the healthcare facility dataset.

\subsection{Explainer: DoctorXAI}\label{ss:drxai}


DoctorXAI by~\cite{panigutti2020doctor} is a post-hoc explainer that can deal with any multi-label sequential model. Since it is agnostic w.r.t. the model, i.e. it does not use any of its internal parameter in the explanation process, it is suitable for our methodology which considers the clinical DSS as a commercial black-box. Furthermore, DoctorXAI exploits medical ontologies in the explanation process, in our case we exploited the ICD-9 ontology. The explanations provided by DoctorXAI are \emph{local} decision rules, which means that they provide the rationale for one particular classification. 

In our scenario, we want to provide an explanation for a over- or under-diagnosis observed in a group of patients, therefore FairLens binarizes the black-box labels and it combines the explanations as described in the \emph{Explanation} and \emph{Reporting} paragraphs of Section \ref{s:pipeline}.

\subsection{Auditing DoctorAI on MIMIC}

\begin{figure}[ht!]
\centering
  \includegraphics[width=\textwidth]{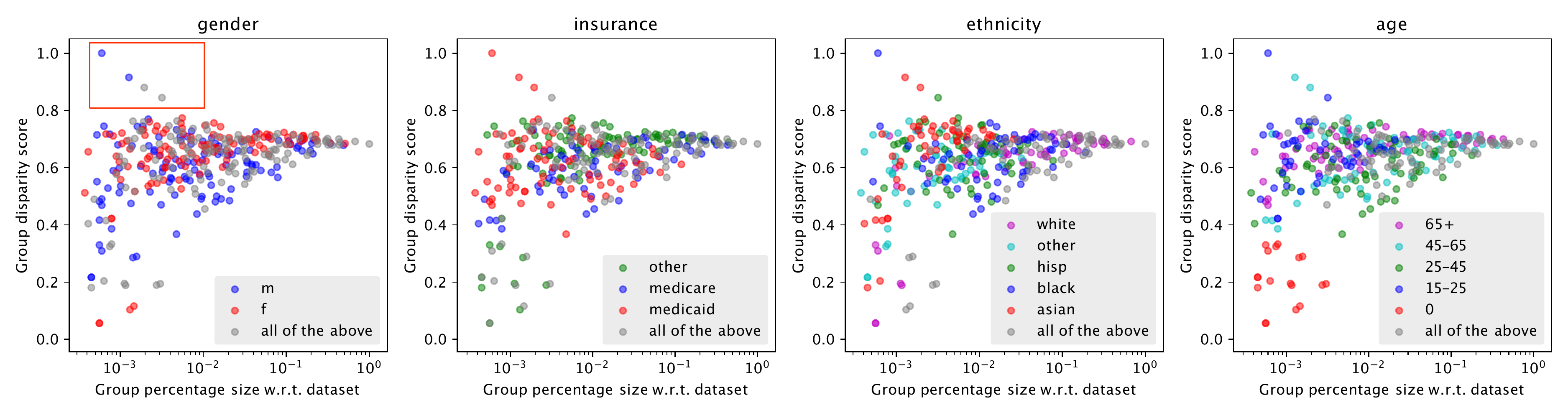}
  \caption{Normalized disparity scores vs. group sizes. In the left, we highlight the top four groups ranked by disparity score.}
  \label{fig:scatter}
\end{figure}

\textbf{Assess the DSS performance on the healthcare facility data} The first step that any domain expert would perform before deploying the clinical DSS on her dataset is to measure its global performance and compare it with the one declared by the manufacturer. In our scenario a domain expert would obtain the results in table \ref{tab:perf}. \\

\begin{table}[h!]
\centering
\begin{small}
\begin{tabular}{llll}
\textbf{BB Recall}      & \textbf{@10} & \textbf{@20} & \textbf{@30} \\ \hline
\textbf{Declared}    & 0.643        & 0.743        & 0.796        \\
\textbf{On auditing data} & 0.481        & 0.623        & 0.712       
\end{tabular}
\vspace*{2mm}
\caption{clinical DSS performance}\label{tab:perf}
\end{small}
\end{table}

\begin{table}[t]
\makebox[\linewidth]{
\begin{scriptsize}
\begin{tabular}{|lcc|lc|lc|}
\cline{1-7}

&&\textbf{Disp.}&\multicolumn{2}{r|}{Over-diagnosed~~~~~~~~~~~~~~\tiny{(Misdiagnosis}}&\multicolumn{2}{r|}{Under-diagnosed~~~~~~~~~~~~~~\tiny{(Misdiagnosis}}\\
\textbf{Group}&\textbf{Size}&\textbf{Score}&&\tiny{Score)}&&\tiny{Score)}\\
\cline{1-7}
%
\multirow{3}{*}{\vtop{\hbox{\strut Male,Medicaid,}\hbox{\strut Black,15-25}}}&\multirow{3}{*}{16}&\multirow{3}{*}{1.0}
&657: Mood disorders& 0.278& 659: Schizophrenia & -0.082\\
&&&98: Essential hypertension& 0.145& 661: Substance-relat. disorders& -0.075\\
&&&651: Anxiety disorders& 0.092& 255: Administrative admission& -0.065\\

\cline{1-7} 

\multirow{3}{*}{\vtop{\hbox{\strut Male,Medicaid,}\hbox{\strut Asian,45-65}}}&\multirow{3}{*}{34}&\multirow{3}{*}{0.915}
&98: Essential hypertension& 0.095& 93: Dizziness& -0.022\\
&&&55: Fluid and electrolyte dis.& 0.047& 157: Acute renal failure& -0.021\\
&&&663: History of mental health& 0.046& 11:  Cancer of head and neck& -0.017\\
\cline{1-7} 

\multirow{3}{*}{\vtop{\hbox{\strut Medicaid,Asian,}\hbox{\strut 45-65}}}&\multirow{3}{*}{52}&\multirow{3}{*}{0.88}
&98: Essential hypertension& 0.095& 93: Dizziness& -0.021\\
&&&259: Residual codes& 0.043& 163: Genitourinary symptoms& -0.019\\
&&&663: History of mental health& 0.04& 102: Chest pain& -0.015\\
\cline{1-7} 

\multirow{3}{*}{Hispanic,15-25}&\multirow{3}{*}{85}&\multirow{3}{*}{0.845}

&196: Other pregnancy...& 0.067& 124: Tonsillitis& -0.024\\
&&&98: Essential hypertension& 0.062& 149: Biliary dx& -0.019\\
&&&259: Residual codes& 0.054& 654: Developmental disorders& -0.018\\

\cline{1-7}
\end{tabular}
\end{scriptsize}
}
\centering
\vspace*{2mm}
\caption{Groups ranked by disparity scores and most over/under-diagnosed conditions when auditing the black-box}
\label{tab:groupaudit}
\label{tab:table}
\end{table}

\textbf{Identify problematic groups of patients}
Once the global performance has been assessed and it is considered good enough, the domain expert can apply FairLens to discover potential biases learned by the model. The domain expert would start by deciding which attributes to use to stratify the patients.For the purpose of our fictional scenario, we consider the following attributes occurring in the auditing data:
%
\begin{itemize}
    \item \emph{Gender:} 54~\% Female, 46~\% Male.
    \item \emph{Ethnicity:} 69~\% White, 18~\% Black, 7~\% Hispanic/Latino, 3~\% Asian, 3~\% Other.
    \item \emph{Age groups:} 0.5~\% infants (age=0), 3.2~\% age 15-25, 18.3~\% age 25-45, 36~\% age 45-65, 42~\% over 65.
    \item \emph{Insurance type:} 41\% Medicare, 12\% Medicaid, 47\% Other.
\end{itemize}
Once these attributes are selected, FairLens computes the disparities across groups. In our scenario, the black-box is a sequential multi-label model that predicts the set of codes diagnosed in the next visit in terms of CCS codes. In this multi-label case, the disparity is evaluated using the \emph{Wasserstein distance}. This metric measures the distance between two probability distributions: for each group of interest, the distance between the CCS codes' distribution in the black-box output and the same distribution in the ground truth. The domain expert can decide to either explore a specific group of interest or to have a comprehensive view of the biases of the DSS on all possible groups. In the latter option, the output of this computation is a set of scatter plots similar to those shown in Figure \ref{fig:scatter}.\\


The scatter plots show the relationship between the group size and its disparity measure. Each point is a combination of attributes, considering all possible combinations of values FairLens obtains 432 groups. The highest variability in terms of disparities is observed in the smaller groups. This would inform the domain expert that fairness issues might be present in relatively rare patients treated in her healthcare facility. On the contrary, different operational strategies should be implemented if the larger disparities were observed over large groups.
The color-coding allows to explore the disparities of each intersectional identity. Data points labelled and color-coded as {\em all of the above} correspond to groups that do not represent a specific value for the stratification feature: for instance, the group {\em (male, medicare)} includes patients of all ages and ethnicities.

The main insights that the user can draw from the scatter plot in this specific case is that the DSS works very well on the infant patients, and there is a tendency of the DSS to misdiagnose patients over 65 years old (Figure~\ref{fig:scatter}.age). Furthermore, Medicaid patients seems to be more frequently misdiagnosed than Medicare patients. \\

\textbf{Identifying systematic sources of error in the selected subgroup}

The domain expert auditing the system can further select a specific group for a more in-depth investigation. Suppose she decides to focus on the groups with the highest disparity (outliers highlighted with a red rectangle in Figure~\ref{fig:scatter}). In this case, FairLens compares the ground truth clinical conditions with the codes predicted by the DSS of these groups, and it isolates the most over- and under-diagnosed CCS codes by the DSS, obtaining the results reported in Table \ref{tab:table} (only the top 4 groups by disparity scores are shown, and only the top/bottom codes ranked by misdiagnosis scores are reported). This analysis tells the domain expert that the group with the highest disparity is young (15-25 y.o.) male black patients with Medicaid. For this group, the clinical DSS tends to over-diagnose \emph{Mood disorders}, \emph{Essential Hypertension} and \emph{Anxiety disorders} while it tends to under-diagnose \emph{Schizophrenia}, \emph{Substance-related disorders} and \emph{Administrative admission}. A similar consideration can be drawn for other patient groups. \\



\begin{figure}[t]
\centering
  \includegraphics[width=0.5\textwidth]{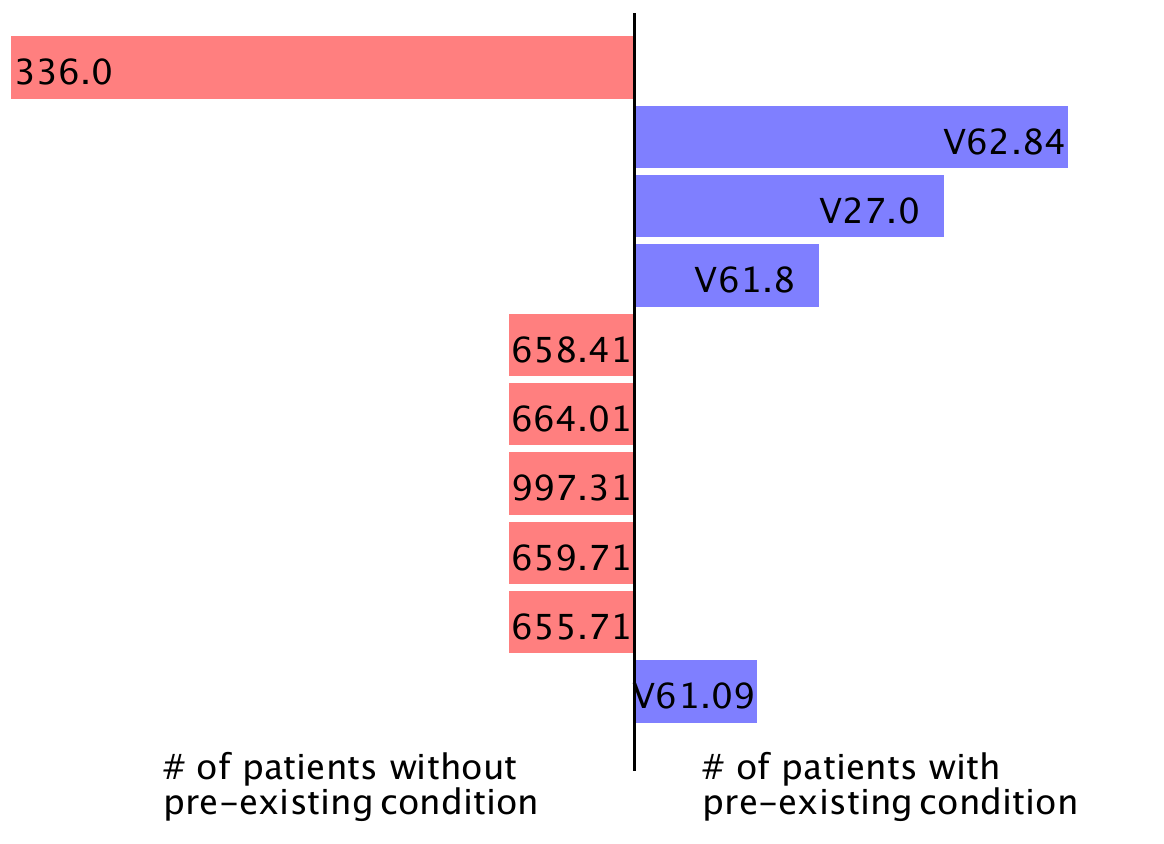}
  \caption{Explanations for misclassified over-diagnosis of Mood disorder.}
  \label{fig:explanations}
\end{figure}

\textbf{Obtaining explanations for systematic misclassifications} Once the groups with the highest disparities are identified, the domain expert can use FairLens to obtain an explanation for one particular misclassification. Consider, for example, the over-diagnosis of \emph{Mood Disorder} (CCS code 657) for young black patients with Medicaid. FairLens uses DoctorXAI to discover which elements in the patients' clinical history drive this over-representation of that specific CCS. This is done by first projecting the black box's multi-label output on the single label 657 (as explained in Section  \ref{s:pipeline}), then calling DoctorXAI to explain the binarized outcome. Highlighting the most common ICD-9 conditions emerging from the group explanations, FairLens shows the domain expert Figure \ref{fig:explanations}. 

The bars' magnitude shows the observed frequency of that code in the group explanations, while the color and the direction of the bar show if the condition associated with that code was its presence or absence in the clinical history.

The explainability report given by FairLens should be read like this: the top 3 ICD-9 codes driving the DDS's over-diagnosis of Mood Disorders in young black patients with Medicaid are \emph{Syringomyelia and Syringobulbia} (ICD-9 code 336.0), \emph{Suicidal ideation} (ICD-9 code V62.84) and \emph{Outcome of delivery, single liveborn} (ICD-9 code V27.0). In particular, the DSS over-diagnoses \emph{Mood Disorders} when \emph{Syringomyelia and Syringobulbia} are not observed in the patient's clinical history and when \emph{Suicidal ideation} and \emph{Outcome of delivery} are present in the patient's clinical history.

%% file: 05_conclusions.tex
\section{Conclusions and Future Work}\label{s:conclusions}

Fairness and explainability are key features to gain trust from patients and clinicians. As black-box ML-based clinical decision support systems will be deployed in real-world healthcare settings, systematic auditing procedures must be in place. 
In this paper we proposed FairLens, an algorithmic pipeline to inspect clinical DSS and we showed how to use it to spot fairness issues in patients' subgroups, and to further investigate potential over-/under-diagnosed conditions. Moreover, the proposed methodology is able to drive domain expert to investigate the reason behind the systematic black-box misclassification by pointing to the most common causes of error within groups through XAI techniques. Future work will be devoted to test FairLens on different black-box and to include additional disparity scores with the aid of domain experts to increase the tool usability. Moreover, we aim at providing a more structured local-to-global explanation \cite{setzu2019global} of the $BB$ error. Finally, additional experiments to generate counterfactual examples will be implemented to increase FairLens adoption from domain experts.